\documentclass{CVIS}
\usepackage{amsmath,bm}
\usepackage{graphicx}
\usepackage{dcolumn}
\usepackage{subcaption,soul}
\usepackage{csvsimple}
\usepackage{caption}
\usepackage[numbers,sort&compress,sectionbib]{natbib}
\usepackage[pagebackref=false,breaklinks=true,letterpaper=true,colorlinks=true,linkcolor = blue, urlcolor = black,citecolor=black,bookmarks=true]{hyperref}
\usepackage{url}
\usepackage{mleftright}
\usepackage{multirow}

\usepackage{titlesec}
\usepackage[redeflists]{IEEEtrantools}

\title{Where Should We Begin? A Low-Level Exploration of Weight Initialization Impact on Quantized Behaviour of Deep Neural Networks}

\begin{document}
\bstctlcite{IEEEexample:BSTcontrol}
\sloppy
\author{
\begin{tabularx}{\textwidth}{X X}
Stone Yun & University of Waterloo\\
Alexander Wong & University of Waterloo\\
\multicolumn{2}{l}{Email: \{s22yun, a28wong\}@uwaterloo.ca}
\end{tabularx}
}

\maketitle
\begin{abstract}
	With the proliferation of deep convolutional neural network (CNN) algorithms for mobile processing, limited precision quantization has become an essential tool for CNN efficiency. Consequently, various works have sought to design fixed precision quantization algorithms and quantization-focused optimization techniques that minimize quantization induced performance degradation. However, there is little concrete understanding of how various CNN design decisions/best practices affect quantized inference behaviour. Weight initialization strategies are often associated with solving issues such as vanishing/exploding gradients but an often-overlooked aspect is their impact on the final trained distributions of each layer. We present an in-depth, fine-grained ablation study of the effect of different weights initializations on the final distributions of weights and activations of different CNN architectures. The fine-grained, layerwise analysis enables us to gain deep insights on how initial weights distributions will affect final accuracy and quantized behaviour. To our best knowledge, we are the first to perform such a low-level, in-depth quantitative analysis of weights initialization and its effect on quantized behaviour.
\end{abstract}

\vspace{-0.1in}
\section{Introduction}
\vspace{-0.05in}

Deep Convolutional Neural Networks (CNN) have enabled dramatic advances in the field of computer vision. An explosion of recent research has demonstrated incredible results in applications such as image classification~\cite{Alexnet, ResNet, Inceptionv4}, object detection~\cite{YOLO, SqueezeDet, Google_SSD, Fast_RCNN}, image segmentation~\cite{UNet, Mask_RCNN} and many more. As CNNs have become an increasing part of everyday life, their usage has expanded to mobile processing. With this expansion comes issues of minimizing latency, area, and power. Fixed point quantization of CNN weights and activations has become an essential tool for running efficient CNN inference. Various works have explored different quantization algorithms~\cite{TFQuantize, LogQuant, VectorQuant} to minimize the loss of information when mapping the weights and activations of a CNN into a discretized space. With these methods, the weights and activations can be represented as an n-bit (most often 8-bit) integer rather than 32-bit floating point numbers. Consequently, simple integer multiply-accumulate (MAC) operations can be performed rather than costly floating point arithmetic. This leads to significant savings in both computation (integer arithmetic) and storage (typically 8-bits per value or less).

Mobile hardware accelerators are usually limited in the types of operations that can be massively parallelized for fast execution. Thus, more complex quantization methods are often not supported by existing hardware. As such, other works have focused on quantization-algorithm-specific optimization methods (Eg. targeting 8-bit uniform quantization). These include quantization-aware fine-tuning \cite{TFQuantize} and differential optimization of quantization parameters \cite{TQT, PACT}, eg. finding the optimal max/min thresholds of each weight/activation tensor for minimal quantized degradation. These methods train a model that is robust to quantized perturbations by simulating the error/noise of fixed point arithmetic.

Despite significant works demonstrating ways to recover near floating-point performance using reduced precision inference, there is still little understanding of how different design decisions can affect the quantized inference behaviour of CNNs. Weights initialization strategies are often designed with the goal of solving issues such as vanishing/exploding gradient \cite{GlorotInit, HeInit, hanin2018}. However, an often-overlooked aspect of weights initialization is its impact on the final trained distributions of each layer. As they determine our starting point on the loss surface, initial distributions of each weight tensor will have a profound impact on the final trained model. Gradient descent is an incremental process with many small, noisy steps. Thus, an intelligent weight initialization strategy will have significant impact on the local minima reached on our path through the loss space. With regards to quantization, this means that weight initialization choices will have significant impact on the dynamic ranges of the weights and activations in a trained CNN. Thus, affecting the noise in our system and the expected quantized inference behaviour.

We propose a framework for in-depth, fine-grained quantitative analysis of the impact of various weights initialization strategies on final accuracy and quantized behaviour. By analyzing the trained distributions of each layer's weights and activations, we can gain deep insight on how different weights initialization strategies will affect the dynamic ranges of each layer. This in turn provides insight on the quantized behaviour of a CNN. Furthermore, we analyze the effect of these different weights initializations for a small set of different CNN architectures. Thus, we are able to isolate and observe the interplay between the CNN architecture choices (the parameterization) and the weights initialization strategy (the starting point on the parameterized loss surface). To our best knowledge, we are the first to perform such a systematic, low-level, quantitative analysis of weights initialization strategies and quantized behaviour. Furthermore, our framework for fine-grained analysis is applicable to analyzing any number of CNN design choices such as layer types, batch size, learning rate schedule etc.

\vspace{-0.1in}
\section{Background}
\vspace{-0.05in}
In early research, neural network parameters were often randomly initialized based on sampling from a normal or uniform distribution. The respective variance and range of these distributions would be hyperparameters for the practitioner to decide. While easily taken for granted, several works such as \cite{GlorotInit, HeInit, hanin2018} have provided rigorous mathematical proofs showing how intelligent weights initialization strategies can solve issues of vanishing and exploding gradients. These works define \textit{fan\_in} and \textit{fan\_out} of a fully connected layer as the input/output units respectively. For convolution, it is defined as Eq.~\ref{eq:fan_inout} where \textit{K} is the kernel width (assume square kernel). They provide mathematical proofs on their proposed fan\_in/fan\_out-aware initialization strategies that scale the variance of gradients at each layer. Thus, avoiding failure modes created by vanishing and exploding gradients. While the introduction of Batch Normalization (BatchNorm) \cite{BatchNorm} layers has greatly mitigated training issues involving gradient scales, the choice of "where to begin" in the parameterized loss space is still extremely relevant. An often-overlooked effect of these initialization strategies is their impact on the trained dynamic ranges of each layer. As gradient descent is a noisy, iterative process with small, incremental steps, the final dynamic ranges of each layer are profoundly impacted by their starting point.
\begin{equation}
\label{eq:fan_inout}
fan_{in/out} = K \times K \times channel_{in/out}\
\vspace{-0.05in}
\end{equation}

\vspace{-0.1in}
\section{Fine-grained Layerwise Analysis}
\vspace{-0.05in}
Besides a high-level study of how different weight initializations affect 32-bit floating point (fp32) and eight-bit quantized (quint8) accuracy, we also wish to gain detailed insight on the layer-wise distributions of final trained weights and activations. This information can give us an in-depth look at how the learning dynamics of various weight initializations play out. Furthermore, the dynamic ranges of each weight/activation tensor determine the resolution of the quantized step-size and, by extension, the quantization noise in a CNN. Thus, this analysis can help explain the observed quantized inference behaviour of different trained models. We propose systematically ablating through a variety of different weight initialization strategies while tracking the dynamic ranges of each layer's weights and activations during training. In this way, we can isolate the effect of these different design choices and analyze the changing distributions at each layer. We also track the "average channel precision". Average channel precision is defined as Eq.~\ref{eq:averageprecision}. Channel precision in this context is the ratio between an individual channel's range and the range of the entire layer. \cite{Nagel_DFQ} uses this precision quantity to algorithmically maximize the channel precisions of each layer in a network prior to quantization. It can be seen as a measure of how well the overall layer-wise quantization encodings represent the information in each channel.
\begin{equation}
\label{eq:averageprecision}
average\_precision = \frac{1}{K}\sum_{i=1}^K \frac{range_i}{range_{tensor}}\
\end{equation}

For dynamic ranges of activations, we randomly sample N training inputs from our training set and observe the corresponding activation responses. To reduce outlier noise, we perform symmetric percentile clipping (Eg. top and bottom 1\%) and track the dynamic range and average precision of the clipped activations. As percentile clipping has become a ubiquitous default quantization setting we feel that this method establishes a realistic baseline of what can be expected during inference-time. Finally, there is one more set of dynamic ranges that must be observed. Batch Normalization has become the best-practice in a large range of CNN algorithms. However, their vanilla form is not well-suited for mobile hardware processing. Best practice for fast CNN inference usually involves folding the scale and variance parameters of a BatchNorm layer into the preceding layer's convolution parameters prior to quantization, as shown in Eq.~\ref{eq:bn_fold}. Therefore, we must also track the dynamic range and precision of our CNN's batchnorm-folded (BN-Fold) weights. In this manner, we can iterate through various weight initializations, gaining insights at each step on the trained models and their learning dynamics as well as the final weights and activations distributions. Our method can be extended to analyze a plethora of other design choices. These can include architecture choices such as layer-type, skip/residual connections as well as training hyperparameters such as learning rate schedules, batch size, optimizers etc. Despite their simplicity, such analyses can provide deep insight on the interplay of these various design choices and perhaps yield new understanding on their interaction.

\begin{equation}
\label{eq:bn_fold}
w_{fold} = \frac{\gamma w}{\sqrt{EMA(\sigma^2_B) + \epsilon}}
\vspace{-0.1in}
\end{equation}

\section{Experiment}
\begin{figure*}
\vspace{-0.6in}
\centerline{\includegraphics[width=13cm]{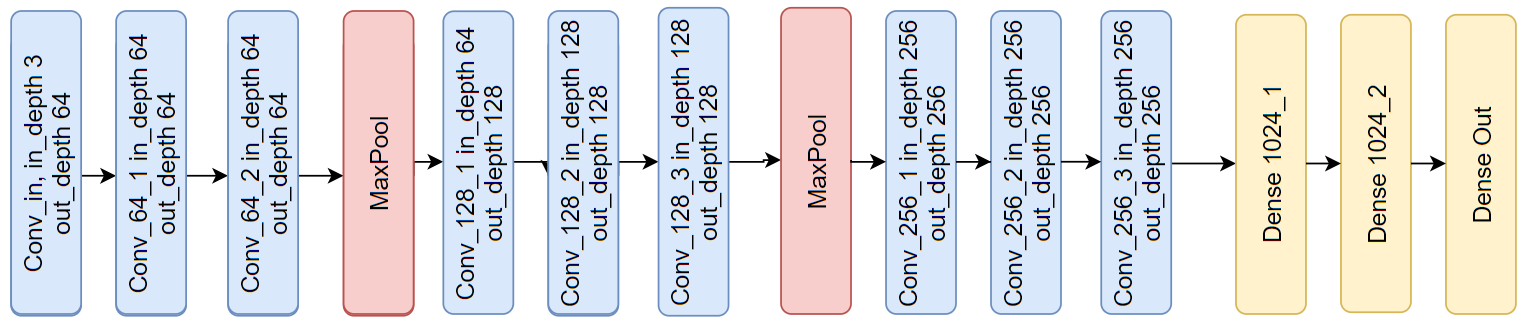}}
\caption{\footnotesize{}\textbf{General Macroarchitecture of the CNN}. For our analysis we use a fixed macro-architecture so that we can isolate the interactions between various weight initialization strategies and a few different convolutional layer choices. We train four variations of this macro-architecture determined by the type of conv-block used at each layer: Regular\_Conv\_With\_BN, Regular\_Conv\_No\_BN, DWS\_Conv\_With\_BN, and DWS\_Conv\_No\_BN. These respectively correspond to using regular convolution followed by BatchNorm and Relu, regular convolution followed by only Relu and no BatchNorm, depthwise separable convolution blocks with BatchNorm and Relu after each convolution layer (same as the MobileNets block in \cite{MobileNet}), and finally depthwise seperable convolution with only Relu and \textbf{no} BatchNorm after each convolution layer.  The very first convolution layer stays fixed for all architectures, but follows the With/Without BatchNorm behaviour of the rest of the layers.}
\vspace{-1.5in}
\label{fig:factorizenet}
\end{figure*}

\vspace{-0.05in}
\label{experiment}
For our experiment we use a simple, VGG-like macroarchitecture with four variations that differ in the micro-architecture of each layer (eg. type of convolution block used, use of BatchNorm and Relu etc. See Figure~\ref{fig:factorizenet} for the general macro-architecture and details on the different variations of convolution layers). Our four CNNs are trained and tested on CIFAR-10 with a wide variety of different weight initialization strategies. These strategies can be separated into two categories of naive, straightforward strategies and more intelligent, layer-aware methods. Furthermore, the most common random weight initializations can also be categorized by the type of sampling distribution: random sampling from uniform distributions (hereafter referred to as RandUni) and random sampling from normal distributions (hereafter referred to as RandNorm). With considerations of dynamic range in mind, we seek to select distributions for the naive methods that would roughly correspond to small, medium, and large initial weights ranges. For the layer-aware initialization strategies, we use four commonly used methods introduced in \cite{GlorotInit, HeInit}. Named after the authors, we call them Glorot Uniform (GlorotUni) and Glorot Normal (GlorotNorm) from \cite{GlorotInit}, He Uniform (HeUni) and He Normal (HeNorm) from \cite{HeInit}. In these works, the distribution range (for uniform sampling) and standard deviation (for normal sampling) for each layer are calculated based on \textit{fan\_in}, \textit{fan\_out}, or some combination of the two. We choose to focus on only the convolution layers and so the fully connected layers are always initialized using Glorot Uniform initialization. Furthermore, we also keep the weight initialization of the first convolution layer constant; only Glorot Uniform initialization is used. This was to keep the very first convolution layer as constant as possible.

Based on initial results showing Glorot Uniform having the most success in fp32 accuracy, we further experiment with Modified Glorot Uniform (ModGlorotUni) weights initialization strategies. The method of computing the max/min range of the uniform sampling distribution in Glorot Uniform initialization can be generalized as Eq.~\ref{eq:glorot_uni}. In the original paper, \textit{C = 6}. Following our established method of selecting distributions corresponding to small, medium, and large initial weights ranges, we select two values of C that would roughly correspond to medium and large ranges. The original Glorot Uniform leads to fairly small ranges. See Table~\ref{weights_init} for a detailed breakdown of the sampling methods used in each of the 48 experiments.

\begin{equation}
\label{eq:glorot_uni}
max/min = \pm \sqrt{\frac{C}{fan\_in + fan\_out}}
\end{equation}

\begin{table}
	\caption{List of the various weight initialization strategies used. For methods that require some hyperparameter selection we include the values selected.}
	\setlength{\tabcolsep}{4pt}
	\renewcommand{\arraystretch}{1.5}
	\centering
	\scalebox{1}{
		\csvreader[tabular=|l|c|c|c|,
        table head=\hline Initialization Method & Standard Deviation & Max/Min Value & C\\\hline,
        late after line=\\\hline]%
        {weights_init_types.csv}{Initialization Method=\Init, Standard Deviation=\Std, Max/Min Value=\MaxMin, C=\C}%
        {\Init & \Std & \MaxMin & \C}%
		}
		\label{weights_init}
		\vspace{-20pt}	
\end{table}

Each network is trained for 200 epochs of SGD with Momentum~=~0.9 and batch-size~=~128. Initial learning rate is 0.01 and we scale it by 0.1 at the 75th, 120th, and 170th epochs. For the activation range tracking we perform top/bottom 1\% clipping computed on a random sample of 1024 training samples. Basic data augmentation includes vertical/horizontal shift, zoom, vertical/horizontal flip and rotation. We use Tensorflow for training and quantizing the weights and activations to quint8 format.

For each network we evaluate testing performance with respect to 4 metrics: fp32 accuracy, quint8 accuracy, quantized mean-squared error (QMSE), and quantized crossentropy (QCE). Results are presented in Table~\ref{quant_result}. QMSE refers to the MSE between the fp32 network outputs and the quint8 network outputs after dequantization. Similarly, QCE measures the cross entropy between the fp32 network outputs and the dequantized quint8 network outputs. While QMSE directly measures how much the quint8 network outputs deviate from the fp32 network, QCE quantifies the difference in the distribution of the network outputs. For classification tasks, the quantized network can predict the same class as the fp32 network, despite deviations in logit values, if the overall shape of the output distribution is similar. Therefore QCE can sometimes be more reflective of differences in quantized behaviour. Additionally, we also observe the percent accuracy degradation (change in accuracy divided by fp32 accuracy) of each network after quantization. Though these quantities often track together, there can be scenarios where a network with more QMSE or QCE actually has less relative quantization degradation from a pure accuracy standpoint. This is likely explained by favourable rounding within the network.

\begin{table*}
	\caption{Detailed results for each combination of weight initialization strategy and CNN architecture. The initialization strategies that suffered from vanishing/exploding gradients are ommitted. DWS\_Conv\_No\_BN\_GlorotUni is kept for illustrative purposes.}
	\setlength{\tabcolsep}{4pt}
	\renewcommand{\arraystretch}{1.5}
	\centering
	\scalebox{1}{
		\csvreader[tabular=|l|c|c|c|c|c|c|,
        table head=\hline Network Architecture & FP32 Accuracy & QUINT8 Accuracy & QMSE & QCE & Percent  Accuracy Decrease\\\hline,
        late after line=\\\hline]%
        {frozen_quant_results.csv}{Network Architecture=\Network, FP32 Accuracy=\FP32, QUINT8 Accuracy=\QUINT8, QMSE=\QMSE, QCE=\QCE, Percent Accuracy Decrease=\Percent}%
        {\Network & \FP32 & \QUINT8 & \QMSE & \QCE & \Percent}%
		}
		\label{quant_result}
\end{table*}

\vspace{-0.1in}
\section{Discussion}
\vspace{-0.05in}
\label{discussion}
We can see in Table~\ref{quant_result} that besides affecting the final FP32 accuracy of a given CNN architecture, the weights initialization strategy also has significant impact on the QUINT8 accuracy.  Particularly worth noting is the markedly improved quantized behaviour in the DWS\_Conv\_With\_BN networks trained using RandUni\_Large initialization. Equally noteworthy is the stark drop in QUINT8 accuracy observed with the DWS\_Conv\_With\_BN networks trained with the HeNorm and HeUni weight initializations and the Regular\_Conv\_With\_BN network trained with ModGlorotUni\_Med initialization. As expected, quantized accuracy usually worsened when BatchNorm layers were introduced. This is often attributed to the increased dynamic ranges/distributional shift introduced by BatchNorm Folding.

\begin{figure}
\vspace{-0.13in}
\centerline{\includegraphics[width=0.3\textwidth]{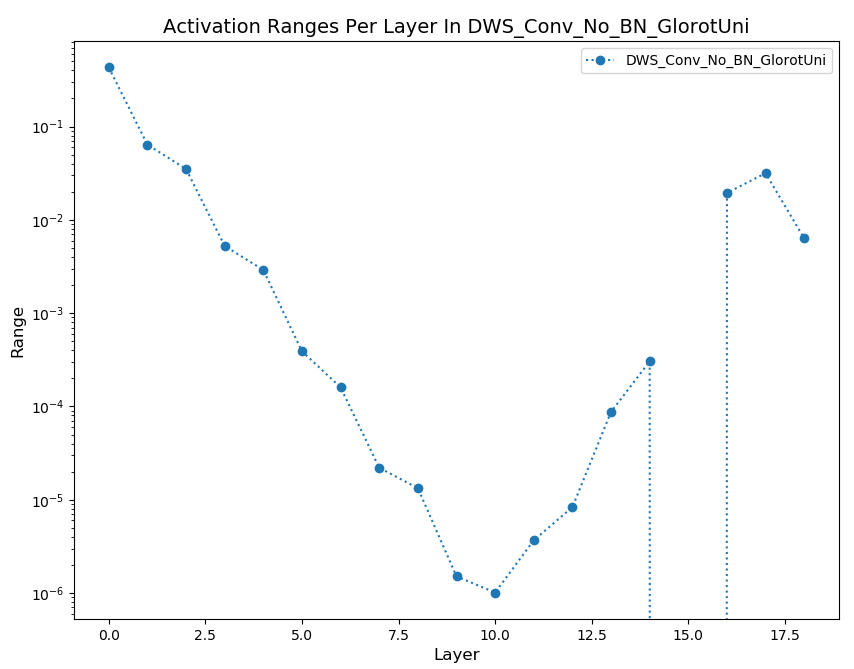}}
\caption{\footnotesize{}\textbf{Vanishing Act!} In this figure we can see how the activation ranges of DWS\_Conv\_No\_BN become increasingly small until they practically disappear. Consequently, gradients are not able to propagate past the fully-connected layers (final three points on the graph).}
\label{fig:vanishing_acts}
\vspace{-0.2in}
\end{figure}

While each CNN architecture is trained on twelve different initialization methods, Regular\_Conv\_No\_BN only has four results. This is because the other initialization methods had issues of exploding gradients. Their results were ommitted. Most of the DWS\_Conv\_No\_BN experiments also did not learn but suffered from vanishing gradient issues instead. However, in our analyses we found that these vanishing gradients were not necessarily caused by a deep architecture leading to the gradient progressively vanishing during backpropagation. Instead, we observed a "vanishing activations" type phenomenon wherein the activations of the final Depthwise Separable Convolution block are exceedingly small. Thus, no gradients are able to propagate past the fully connected layers. Figure~\ref{fig:vanishing_acts} shows a plot of the network activations in DWS\_Conv\_No\_BN\_GlorotUni. For illustrative purposes, we keep the DWS\_Conv\_No\_BN\_GlorotUni result and omit the rest. The normalization introduced by BatchNorm alleviates this issue as expected. One could consider an additional interpretation of BatchNorm as adding capacity to the network in the form of a learned \textbf{explicit scaling}. Scaling that would otherwise be too difficult for the convolution parameters to learn in addition to extracting features. We seek to follow-up on this hypothesis in future works. While we focus on the variations in quantized behaviour in this work, the varying FP32 accuracies are also worthy of close study. Our method sets out a framework through which we can systematically study these phenomena.

\begin{figure}
\vspace{-0.15in}
    \centering
    \begin{minipage}{0.33\textwidth}
        \centering
        \includegraphics[width=0.98\textwidth]{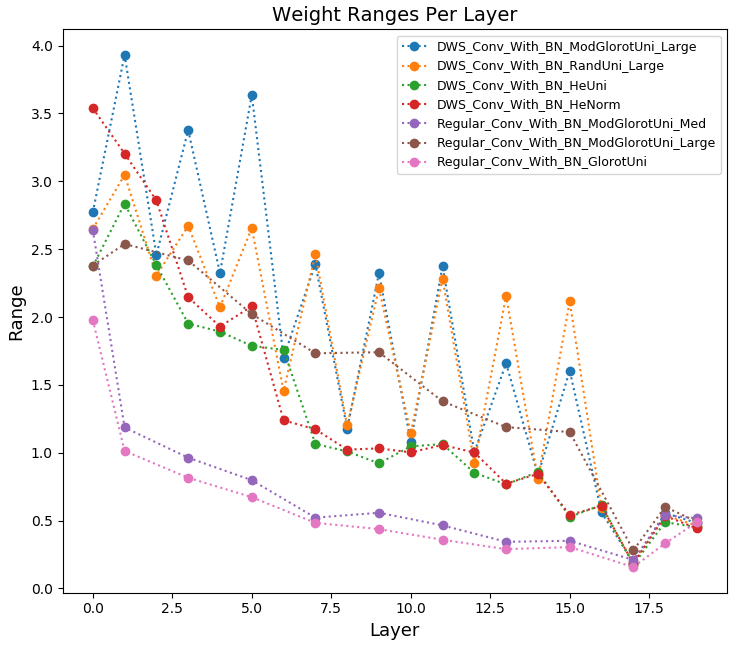} 
    \end{minipage}\hfill
    \begin{minipage}{0.33\textwidth}
        \centering
        \includegraphics[width=0.96\textwidth]{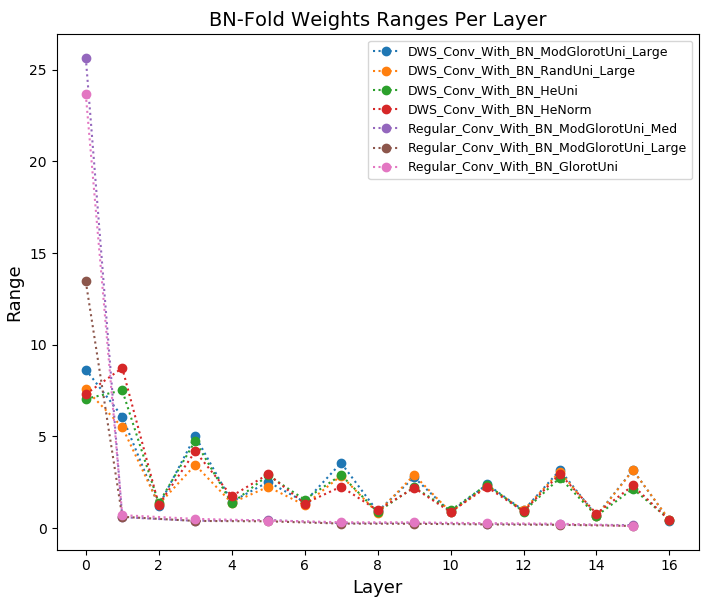} 
    \end{minipage}\hfill
    \begin{minipage}{0.33\textwidth}
        \centering
        \includegraphics[width=0.98\textwidth]{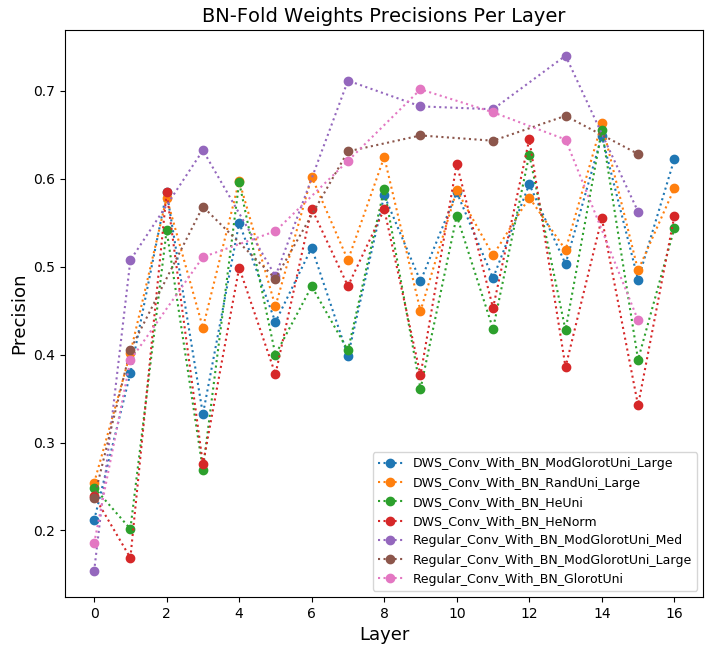} 
    \end{minipage}\hfill
    \begin{minipage}{0.33\textwidth}
        \centering
        \includegraphics[width=0.96\textwidth]{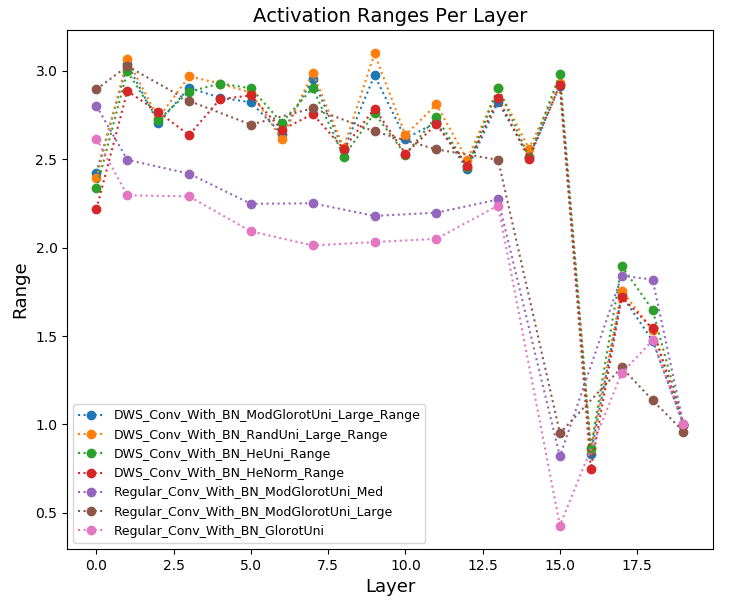} 
    \end{minipage}\hfill
    \begin{minipage}{0.33\textwidth}
        \centering
        \includegraphics[width=0.96\textwidth]{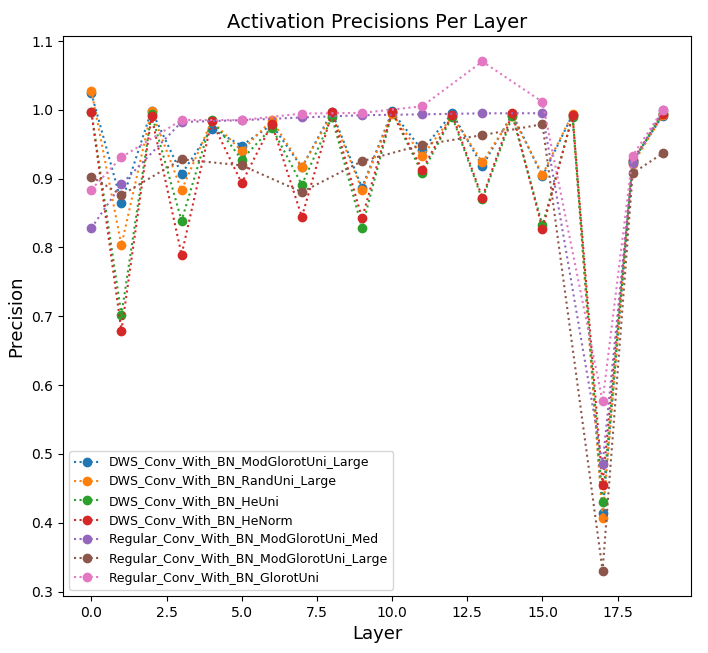} 
    \end{minipage}
    \vspace{-0.07in}
    \caption{\footnotesize{} A more low-level, focused look. Directly comparing a subset of the architectures. In order from top to bottom, the plots show: weights ranges per layer, BN-Fold weights ranges, BN-Fold weights precisions, activations ranges, and activation precisions. As we look to analyze any anomalies or unexpected behaviour, our fine-grained approach allows us to gain much more detailed insight as to what dynamics are at play when we introduce quantization noise.}
    \vspace{-0.22in}
\label{fig:focus-analysis}
\end{figure}

To better understand why we are observing the given quantized behaviour, we can use the proposed fine-grained analysis and inspect the distributions of each model layer-by-layer. With regards to the significantly improved quantized accuracy for DWS\_Conv\_With\_BN\_RandUni\_Large, we observe in Figure~\ref{fig:focus-analysis}~(top) that weights ranges don't necessarily tell the whole story. Despite having generally larger weights ranges, we start to see several other key areas in which the RandUni\_Large layers stand out. For example, while the two He-initialized models tend to have a spike in the BN-Fold weights range at layer 2, RandUni\_Large actually decreases in range. Furthermore, when we compare the BN-Fold weights precisions we also see a drop in precision for the other networks at layer 2 while the precision for RandUni\_Large increases. With the activations, we see that all of the activation ranges increase at layer 2 while activation precisions decrease. However, RandUni\_Large experiences a significantly smaller drop in activation precision. Thus, suggesting that RandUni\_Large has a much higher retention of information in those crucial early stages of low-level feature extraction. Analyzing the change in the layerwise distributions during training might explain \textit{why} we observe such a wide range of behaviour caused by varying weight initialization. It would also be worthwhile to observe the relative change in range/precision after BatchNorm folding. This would be a proxy for observing the distributional shift of the weights. While it is intractable to pinpoint any single reason, our layer-level analysis reveals a rich set of interactions that slowly build a detailed picture of each network’s system dynamics as well as inter-network trends. We could further expand our analysis to use more rigorous, yet scalable statistical methods of analysis. For example, we know that a uniformly distributed tensor would best utilize the quantized steps of our given discretization method. Thus, computing the KL-divergence between a given weight/activation tensor and its corresponding uniform distribution (ie. a uniform distribution with the same bounds as the tensor) is a potential metric to explore. Overall, from these initial analyses, we see that taking a fine-grained, systematic approach to analyzing various design choices can yield detailed insights on the learning dynamics of a CNN. 

\vspace{-0.05in}
\section{Conclusion}
\vspace{-0.05in}
We conduct the first in-depth, quantitative study of the impact of weight initialization strategies on final quantized inference behaviour of various basic CNN architectures.  We show that in addition to affecting final floating point accuracy, a well-chosen weight initialization can also significantly affect a CNN's quantized accuracy. Future work includes  further exploration of the interaction of BatchNorm with initial weight distributions, analysis of other intelligent initialization strategies, and analysis of weight initialization's impact on more complex architectures.

\bibliographystyle{IEEEtran}
\bibliography{references.bib}
\end{document}